# MODEL-BASED INFLUENCE DIAGRAMS FOR MACHINE VISION

T. S. Levitt,* J. M. Agosta,+
T. O. Binford+

*Advanced Decision Systems
Mountain View, California

+Stanford University
Stanford, California

## Abstract

We show the soundness of automated control of machine vision systems based on incremental creation and evaluation of a particular family of influence diagrams that represent hypotheses of imagery interpretation and possible subsequent processing decisions. In our approach, model-based machine vision techniques are integrated with hierarchical Bayesian inference to provide a framework for representing and matching instances of objects and relationships in imagery, and for accruing probabilities to rank order conflicting scene interpretations. We extend a result of Tatman and Shachter to show that the sequence of processing decisions derived from evaluating the diagrams at each stage is the same as the sequence that would have been derived by evaluating the final influence diagram that contains all random variables created during the run of the vision system.

## I. Introduction

Levitt and Binford [Levitt et al.-88], [Binford et al.-87], presented an approach to performing automated visual interpretation from imagery. The objective is to infer the content and structure of visual scenes of physical objects and their relationships. Inference for machine vision is an errorful process because the evidence provided in an image does not map in a one to one fashion into the space of possible object models. Evidence in support or denial of a given object is always partial and sometimes incorrect due to obscuration, occlusion, noise and/or compounding of errorful interpretation algorithms. On the other hand, there is typically an abundance of evidence [Lowe-86].

In our approach, three dimensional model-based machine vision techniques are integrated with hierarchical Bayesian inference to provide a framework for representing and matching instances of objects and relationships in imagery, and for accruing probabilities to rank order conflicting scene interpretations. In particular, the system design approach uses probabilistic inference as a fundamental, integrated methodology in a system for reasoning with geometry, material and sensor modeling.

Our objective is to be capable of interpreting observed objects using a very large visual memory of object models. Nevatia [Nevatia-74] demonstrated efficient hypothesis generation, selecting subclasses of similar objects from a structured visual memory by shape indexing using coarse, stick-figure, structural descriptions. Ettinger [Ettinger-88] has demonstrated the reduction in processing complexity available from hierarchical model-based search and matching. In hierarchical vision system representation, objects are recursively broken up into sub-parts. The geometric and functional relations between sub-parts in turn define objects that they comprise. Taken together, the models form an interlocking network of orthogonal part-of and is-a hierarchies.

Besides their shape, geometrical decomposition, material and surface markings, in our approach, object models hold knowledge about the image processing and aggregation operations that can be used to gather evidence supporting or denying their existence in imagery. Thus, relations or constraints between object sub-parts, such as the angle at which two geometric primitives meet in forming a joint in a plumbing fixture, are modeled explicitly as procedures that are attached to the node in the model to represent the relation. Thus model nodes index into executable actions representing image evidence gathering operations, im-age feature aggregation procedures, and 3D volume from 2D surface inference.

In Binford and Levitt's previous work, the model structuring was guided by the desire to achieve the conditional independence between multiple children (i.e., sub-parts) of the same parent (super-part, or mechanical joint). This structuring allowed Pearl's parallel probability propagation algorithm [Pearl-86] to be applied. Similarly, the concept of value of information was applied to hierarchical object models to enable a partially parallelized algorithm for decision-theoretic system control. That is, the Bayes net was incrementally built by searching the model space to match evidence extracted from imagery. At each cycle, the model space dictated what evidence gathering or net-instantiating actions could be taken, and a decision theoretic model was used to choose the best set of actions to execute.

However, the requirement to force conditional independence may lead to poor approximations to reality in object modeling, [Agosta-88]. Further, the authors did not prove the coherence or optimality of the decision making process that guided system control.

In this paper we make first steps toward formalizing the approach developed by Binford and Levitt. We set up the problem in an influence diagram framework in order to use their underlying theory in the formalization. Image processing evidence, feature aggregation operations used to generate hypotheses about imagery interpretation, and the hypotheses themselves are represented in the influence diagram formalism. We want to capture the processes of searching a model database to choose system processing actions that aggregate (i.e., generate higher level object hypotheses from lower level ones), search (i.e., predict and look elsewhere in an image for object parts based on what has already been observed) and refine (i.e., gather more evidence in support or denial of instantiated hypotheses).

The behavior of machine vision system processing is represented as dynamic, incremental creation of influence diagrams. Matches of image evidence and inferences against object models are used to direct the creation of new random variables representing hypotheses of additional details of imagery interpretation. Dynamic instantiation of hypotheses are formally realized as a sequence of influence diagrams, each of whose random variables and influence relations is a superset of the previous. The optimal system control can be viewed as the optimal policy for decision making based on the diagram that is the "limit" of the sequence.

We extend a result of Tatman and Shachter [Tatman-86], [Tatman and Shachter-89] to show that the sequence of processing decisions derived from evaluating the diagrams at each stage is the same as the sequence that would have been derived by evaluating the final influence diagram that contains all random variables created during the run of the vision system.

In the following we first review our approach to inference, section 2, and control, section 3, in computer vision. In section 4 we represent results of the basic image understanding strategies of aggregation, search and refinement in influence diagram formalisms. In section 5 we sketch a proof of the soundness of control of a vision system by incremental creation and evaluation of influence diagrams.

## II. Model-Based Reasoning for Machine Vision

We take the point of view that machine vision is the process of predicting and accumulating evidence in support or denial of runtime generated hypotheses of instances of a priori models of physical objects and their photometric, geometric, and functional relationships. Therefore, in our approach, any machine vision system architecture must include a database of models of objects and relationships, methods for acquiring evidence of instances of models occuring in the world, and techniques for matching that evidence against the models to arrive at interpretations of the imaged world. Basic image evidence for objects and relationships includes structures extracted from images such as edges, vertices and regions. In non-ranging imagery, these are one or two dimensional structures. Physical objects, on the other hand, are three dimensional. The inference process from image evidence to 3D interpretation of an imaged scene tends to break up into a natural hierarchy of representation and processing, [Binford-80].

Processing in a machine vision system has two basic components: image processing to transform the image data to other representations that are believed to have significance for interpretation; and aggregation operations over the processed data to generate the relations that are the basis for interpretation. For example, we might run an edge operator on an image to transform the data into a form where imaged object boundaries are likely to have high values, while interior surfaces of objects are likely to have low values. We then threshold and run an edge linking operator on this edge image (another image processing operator) to produce a representation where connected sets of pixels are likely to be object boundaries. Now we search for pairs of edges that are roughly parallel and an appropriate distance apart as candidates for the opposite sides of the projected image of an object we have modeled. This search "aggregates" the boundaries into pairs that may have significance for object recognition.

Aggregation and segmentation operations are fundamental in data reduction. We show how the concept of aggregation in bottom up reasoning can be the basis for generating hypotheses of object existence and type. Aggregation applies constraints from our understanding of geometry and image formation. The aggregation operators also correspond to the transformations between levels in the object recognition hierarchies; Sub-parts are grouped together at one level by relationships that correspond to a single node at the next higher level. Therefore grouping operators dictate the "out-degree" of a hypothesis at one hierarchy level with its children at the level below.

Control of a machine vision system consists of selecting and executing image processing and grouping operations, searching the object model network to match groups to models, instantiating hypotheses of possible observed objects or object parts, accruing the evidence to infer image interpretations, and deciding when interpretation is sufficient or complete.

## III. Sequential Control for Machine Vision Inference

Presented with an image, the first task for a machine vision system is to run some basic image processing and aggregation operators to obtain evidence that can be used to find local areas of the image where objects may be present. This initial figure-from-ground reasoning can be viewed as bottom-up model matching to models that are at the coarsest level of the is-a hierarchy, i.e., the "object/not-object" level. Having initialized the processing on this image, basic hypotheses, such as "surface/not-surface" can be instantiated by matching surface models.

After initialization, a method of sequential control for machine vision is as follows:

0. Check to see if we are done. If not, continue.

1. Create a list of all executable evidence gathering and aggregation actions by concatenating the actions listed in each model node that correspond to an instantiated hypothesis.

2. Select an action to execute.

3. Action execution results in either new hypotheses being instantiated, or more evidence being attached to an existing hypothesis.

4. Propagate evidence to accomplish inference for image interpretation, and go to (0).

From our model-based point of view, an action associated with a model node that corresponds to an instantiated hypothesis has one of the following effects: refining, searching or aggregation. In the following we explain these actions. In the next section, we show a method of representing the effects of these actions in an influence diagram formalism.

Refining a hypothesis is either gathering more evidence in direct support of it by searching for sub-parts or relationships on the part-of hierarchy below the model corresponding to the hypothesis, or instantiating multiple competing hypotheses at a finer level of the is-a hierarchy that are refined interpretations of the hypothesized object. For example, given a hypothesized screwdriver handle, in refinement we might look for grooves in the hypothesized screwdriver handle.

Searching from a hypothesis is both predicting the location of other object parts or relationships on the same hierarchy level, and executing procedures to gather evidence in support or denial of their existence. In searching for the screwdriver handle, we might look for the blade of the screwdriver, predicting it to be affixed to one end or othe other of the handle.

Aggregation corresponds to moving up the part-of hierarchy to instantiate hypotheses that include the current hypothesis as a sub-part or sub-relationship. Having hypothesized the screwdriver handle and the screw-driver blade, we can aggregate sub-parts to hypothesize the existence of the whole screwdriver.

In summary, as we spawn hypotheses dynamically at runtime, hypothesis instantiation is guided by a priori models of objects, the evidence of their components, and their relationships. System control alternates between examination of instantiated hypotheses, comparing them against models, and choosing what actions to take to grow the instantiated hypothesis space, which is equivalent to seeing more structure in the world. The possible actions are also stored in the model space either explicitly as lists of functions that gather evidence (e.g., infer-specularity, find-edges, etc.) or that aggregate object components or other evidence nodes. Thus, inference proceeds by choosing actions from the model space that create new hypotheses and relationships between them. It follows that all possible chains of inference that the system can perform are implicitly specified a priori in the model-base.

This feature clearly distinguishes inference from control. Control chooses actions and allocates them over available processors, and returns results to the inference process. Inference uses the existing hypothesis space, the current results of actions (i.e., collected evidence) generates hypotheses and relationships, propagates probabilities, and accumulates the selectable actions for examination by control. In this approach, it is impossible for the system to reason circularly, as all instantiated chains of inference must be supported by evidence in a manner consistent with the model-base.

## IV. Model Guided Influence Diagram Construction

The influence diagram formalism with which we build the model-base allows three kinds of nodes; probability nodes, value nodes and decision nodes. Probability nodes are the same as in belief nets [Pearl-86]. Value nodes and decision nodes represent the value and decision functions from which a sequential stochastic decision procedure may be constructed. The diagram consists of a network showing the relations among the nodes. Solution techniques exist to solve for the decision functions, (the optimal policies) given a complete diagram. Formulating the model-base as an influence diagram allows existing solution techniques [Shachter-86] to be exploited for evaluation of the interpretation process.

The step of generating new hypotheses dynamically upward, from the evidence and hypothesis at the current stage, adds structure to the influence diagram. Expanding the network then re-evaluating it introduces a new operation that is not equivalent to any evalua-tion step for influence diagrams. In a aggregation step, a hypothesis is created to represent a part composed of a set of sub-parts at the lower level. For example, in the domain of low level image constructs, such as lines and vertices, aggregation by higher level parts determines a segmentation of the areas of the image into projected surfaces. This concept of segmentation differs from "segmentation" used in image processing in several ways. First, a common process of aggregation is used throughout the part-of hierarchy; there is no unique segmentation operator. Second, the segmentation need not be complete; the aggregation operator may only distinguish the most salient features. The notion of segmentation as "partitioning a region into segments" no longer applies. Finally, because the refinement step allows the prediction by higher level hypotheses of lower level features that have not yet been hypothesized, the segmentation may be extended by interpretations from above.

Hypothesis generation is implemented by aggregation operators. The combinations of all features at a level by all aggregation operators that apply, is a combinatorially demanding step. To avoid this complexity the adjacency of features is exploited. Features that are aggregated belong to objects that are connected in space. This does not necessarily mean that the features appear next to each other in the image, rather they are near each other in object space. Exploiting this constraint limits the hypotheses generated to a small number of all possible sets of features.

Aggregation operators are derived from the models of parts in terms of the measured parameters of their sub-parts. From a physical model of the part, a functional relation among parameters is derived that distinguishes the presence of the part. In general, the aggregation operator calculates a score, based on distance and "congruence" between a part's sub-parts. Aggregation hypotheses may be sorted so that "coarse" sub-parts are considered before "fine," to further restrict the set of hypotheses generated. As described, this score is a deterministic function of the parameters of

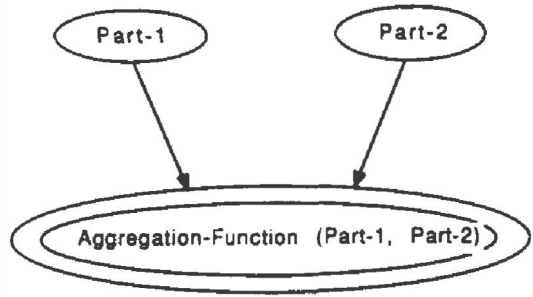

Figure 1: Deterministic Aggregation Process

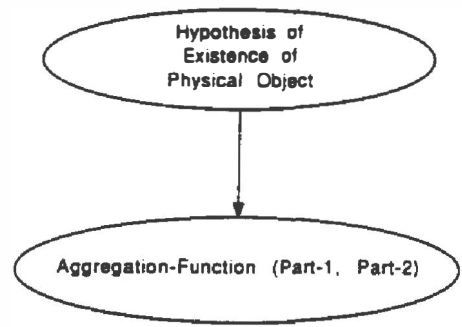

Figure 2: Hypothesis Generation from the Aggregation Process

the features to be aggregated. (See Figure 1.)

The distribution of the aggregation function is conditioned by the hypothesis. It is described by a likelihood, $p\{s|h\}$, the probability of the score, given the hypothesis. (Figure 2.)

From the a model of the appearance of the object, a stochastic model of the distribution of the aggregation score can be derived for the cases that the hypothesis does or does not exist. This likelihood distribution is the probabilistic aspect of the aggregation node, that allows the hypothesis probability to be inferred from the sub-part parameters.

This formulation is valuable because it shows how the the recognition process may be formalized as distributions within a probability net. Consider a search for projected-surface boundaries, to identify the surfaces that compose them. In this instance, suppose the projected-surface boundaries are adjacent parallel lines. To aggregate projected-surface boundaries we derive a scoring function based on both the parallelism and proximity of line boundaries. In searching for projected-surface boundaries, the model generation may disregard most potential boundaries of lines by physical arguments without resort to calculating the aggregation function. Those boundaries for which the scoring rule succeeds spawn a parent node containing a surface hypothesis. This is how the aggregation operator participates in the aggregation process.

A sub-part may be be a member of the sets of several aggregation operators. Further rules are then applied to determine whether hypotheses so formed exclude each other, are independent or are necessarily co-incident. The range of exclusion through co-incidence may be captured in the derivation of the likelihood distributions of a sub-part as it is conditioned on more than one hypothesis.

In general, the diagrams, Figures 1 and 2 are solved by first substituting in the deterministic scoring functions, then applying Bayes rule. To derive a general form for the aggregation operator influence diagram, imagine the aggregation operator as a parent to the part nodes. In Pearl's solution method, the parent receives a lambda message that are functions of the parameters in each of the sub-part nodes. This message contains the aggregation function. Because the aggregation operator expresses a relation among the parts, it may not be factorizable as it would be if the sub-part nodes were conditionally independent, hence the dependency expressed by the aggregation node among the part nodes.

If we consider the aggregation node's clique to involve both the high level hypothesis and the sub-part nodes, then an additional set of arcs appear from the hypothesis to its sub-parts. This is clear when Bayes' rule is written out for the posterior distribution of the hypothesis:

$$p\{h|s\, l_1\, l_2\} = \frac{p\{s|h\}p\{l_1|s\, h\}p\{l_2|s\, h\, l_1\}p\{h\}}{p\{s\, l_2\, l_1\}}$$

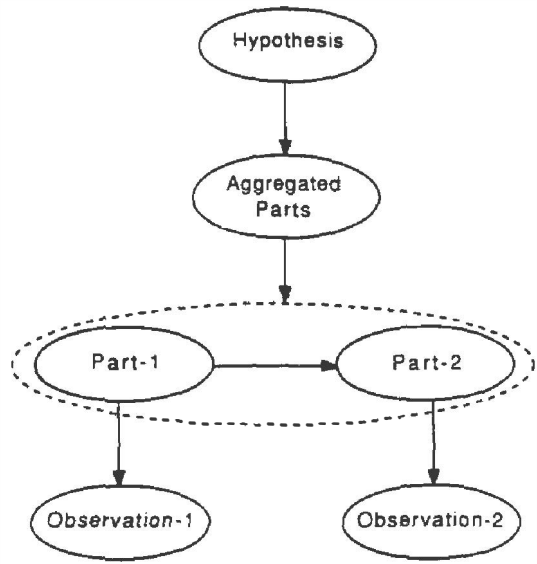

Figure 3: Generic Influence Diagram Probability Nodes for Machine Vision

The aggregation operator likelihood appears multiplied by a set of other factors. The additional terms like $p\{l|h\}$ we term "existence" likelihoods. They are the arcs to the sub-parts, $l_i$, from the hypothesis, $h$. Their interpretation is, given $h$ is observed (or is not observable) does the sub-part appear? Most often these are certainty relations: If there is no obscuration, existence of $h$ implies appearance of its composite features, and vice versa. Thus they may express observability relations where $h$ exists but not all of its features are observed.

To further clarify, think of each feature node's state space as the range of parameters that describe it, plus one point – that the node is not observed. The probability that the node appears is the integral of all the probability mass over the range of parameters. Thus each part can be envisioned as two probabilistic nodes; one a dichotomy, either the part is known to exist or it is not; the other a distribution over parameters that describe the location and shape, dependent on the existence node. The aggregation function expresses a relation between composite sub-part parameters and the existence of the parent. The additional terms in Bayes rule suggests di-rect relations between the existence node of the parent and appearance sub-part features. These additional terms may be thought of as the membership relations in the is-part-of network. The relation between the parameters of the sub-parts and the parent's *parameters* poses an additional inference problem, much along the lines of traditional statistical inference of estimating a set of model parameters from uncertain data.

This method emphasizes the use of measured and inferred values to determine the existence of features; we are converting parameters into existence probabilities as we move up the network. The method concentrates on the classification aspect rather than the estimation and localization aspect. The hope is that once a set of stable, high level hypotheses are generated, the more difficult part of recognition has been solved, and accurate estimation can follow using the data classification generated by what is effectively an "interpretation driven" segmentation process. Estimation can be thought of as a "value to value" process. It might well be necessary to carry this out concurrently if accurate values are required. Alternately, evidence may enter the network directly at higher levels. Neither possibility presents a problem to the algorithm.

## V. Dynamic Instantiation for Sequential Control

In this section, we present a way to formalize the control problem for inference up the machine vision hierarchy. We show how control over the hierarchy can be expressed as a dynamic program by an influence diagram formulation. At this level of generality we can abstract out the structure at each level and coalesce all hypotheses at one level of the hierarchy into one node. These hypotheses nodes form a chain from the top level (the object) hypothesis to the lowest level. Each level has corresponding aggregation and, possibly, evidence nodes for the aggregation process at that level. This high level structure lets us show that for purposes of control the level of the hierarchy can be considered as stages of a dynamic programming problem. Thus each level has the structure shown in Figure 3.

Each stage in the dynamic program is constructed from the aggregation operators at one level of the hierarchy. We add decision and sub-value nodes to the influence diagram to represent control in a dynamic program. In the following, we use $e_i$ to represent the $i$-th set of observations (i.e., evidence from image processing operators), $g_i$ to represent the $i$-th aggregation score, $h_i$ to represent hypotheses about physical objects, $d_i$ to represent processing decisions, and $v_i$ to represent control costs. The $V$ node represents the values assigned to the top-level hypotheses.

The process starts at the bottom of the diagram with the first aggregation forming the first set of hypotheses from the original evidence. The evidence may guide the choice of aggregations; which we show by the decision, $d_0$, with a knowledge arc from $e_0$. An example would be to choose an edgel linking aggregation operator as $g_1$, where $e_0$ are edgels found in an image, and $h_1$ are hypothesised object boundaries. This first stage is shown in Figure 5.

The final decision, $d_1$, selects the object hypothesis with the highest value. It will float to the top stage as we add more stages. The top level value $V$ depends on the object hypotheses. Intermediate hypotheses do not contribute to the value; stage decisions only affect the costs of calculation, $v_i$, which are additive, as the dynamic programming formulation requires. It may be interesting to consider what are the computational gains from a value function that is separable by object hypotheses; such a value function is not considered here.

Next the system makes a decision of which processing action to take at the superior stage. If we add the decision at $d_1$ to, for example, match boundaries into parallel sets with aggregation operator $g_2$ and so generate projected-surface hypotheses, $h_2$, we have the diagram shown in Figure 6. Here $d_2$ is, as described, the choice-of-object decision.

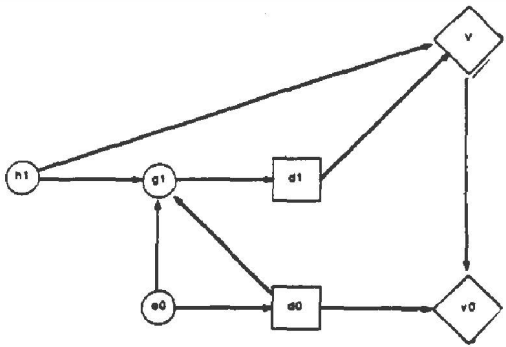

Figure 4: First Inference Stage Influence Diagram

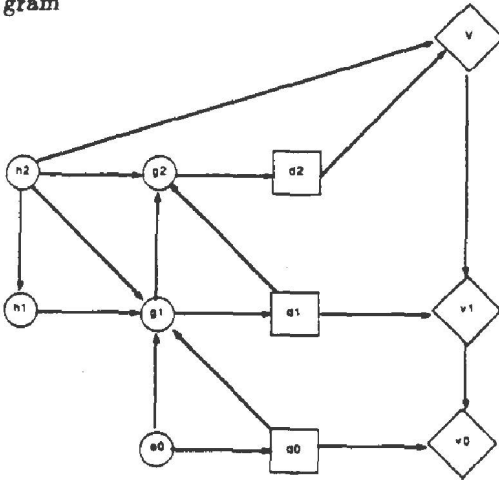

Figure 5: Aggregation Processing Second Level Influence Diagram

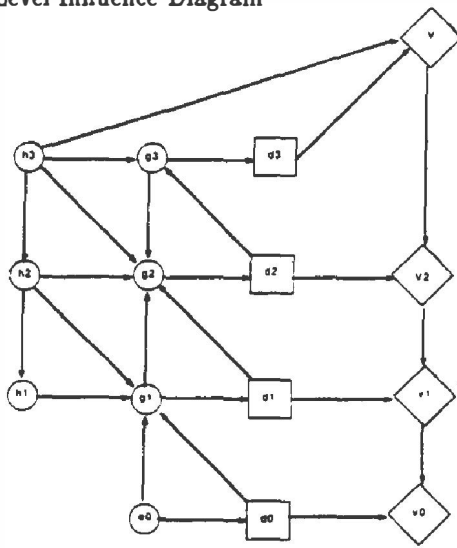

Figure 6: Aggregation Processing Third Level Influence Diagram

We can continue to iterate the diagram building process to add another aggregation stage, as shown in Figure 7. It is clear how the sequence of diagrams proceeds as we continue to generalize upward to complete the part-of hierarchy. If we look at the sequence of influence diagrams from initialization to object recognition, then we can regard the final diagram as if it had been built before evaluation took place. The distributions within the nodes will differ depending on the solution to the diagram. It follows that if we show that the evaluation method is sound in terms of legal influence diagram operations, then we have a formal framework with which to develop an optimal recognition scheme, and in particular a value based method of control.

These results are an application of work by Tatman and Shachter [Tatman-85], [Tatman and Shachter-89] on sub-value nodes and dynamic programming techniques represented in influence diagram form. Tatman shows that optimal policies for diagrams such as those above can be obtained by influence diagram techniques that are equivalent to dynamic programming methods, and like these methods increase linearly in complexity with the number of stages.

In particular, Tatman's influence diagram realization of Bellman's Principle of Optimality [Bellman-57] states that in a diagram with stage decision variables $d_1, ..., d_n$, if there exists a set of nodes $X_k = \{x_k(1), x_k(2), ...\}$ associated to each decision $d_k$, such that

1. all nodes in $X_k$ are informational predecessors of $d_k$

2. the value node is a sum or product of sub value nodes

3. at least one element of $X_k$ is on every directed path from the predecessors of $X_k$ to the successors of $X_k$ (with the exception of the value nodes), then the optimal policy for the decision process, $\{d_1*, ..d_n*\}$, will have the property that policy $\{d_k*, d_{k+1}*, ...d_n*\}$ is optimal for the decision process defined by the original decision process with all nodes, except $X_k$ and its successors deleted.

If at each level in the hierarchy we set the aggregation node, $g_k$ equal to the set $X_k$, we have met the requirements of Tatman's Theorem.

So far our influence diagram does not allow the incorporation of evidence above the lowest level. Decisions above $d_1$ receive no evidence in addition to the deterministic aggregation computation from the level below. We now consider the representation of the processes of search and refinement. These operations will extend the range of actions at a decision node to incorporate evidence hypotheses at the same level or just above.

Notice that aggregation is the process of generating a hypothesis at the next higher level. As such it is a process of generalization from sub-parts to hypothesized super-parts. In comparison, search is a process of adding more evidence to dis-ambiguate the competing higher level hypotheses. This is typically done by using the object models in combination with the location of currently hypothesized imaged objects to direct search and processing elsewhere in the image. For example, having hypothesized a projected-surface, $h_2$, we could search in the region bounded by the projected-surface boundaries (from $g_2$) to run a region operator to infer surface-like qualities, or we could search near the projected-surface to attempt to infer neighboring surfaces. The influence diagram structure is pictured below.

Either operation involves gathering more imagery-based evidence. Hence we denote this as $e_3$, because it corresponds to the third level of the processing hierarchy. Notice that there is no direct dependency between $g_3$ and $e_3$. By letting Tatman's $X_3 = \{g_3, e_3\}$, we still fulfill the requirements of Bellman's theorem.

We now turn to refinement. A refinement operation might be to run an operator over the projected-surface, $h_2$, that compared contrasts across the projected-surface boundaries to see if they were likely to bound the same projected surface. Such an operator is chosen after $h_2$ is instantiated, and so is made at decision time $d_2$. In this way the evidence collected revises a hypothesis already "aggregated." This is the critical distinction between refinement and search. We view it as providing additional evidence about $h_2$, so we call it $e_2$; $e_2$ "refines" the hypothesis $h_2$. This process is pictured below.

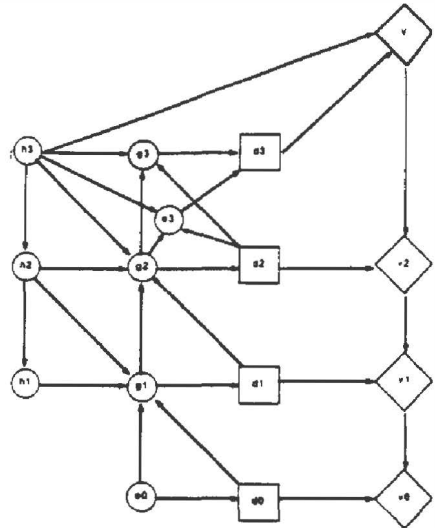

Figure 7: Search Process Influence Diagram

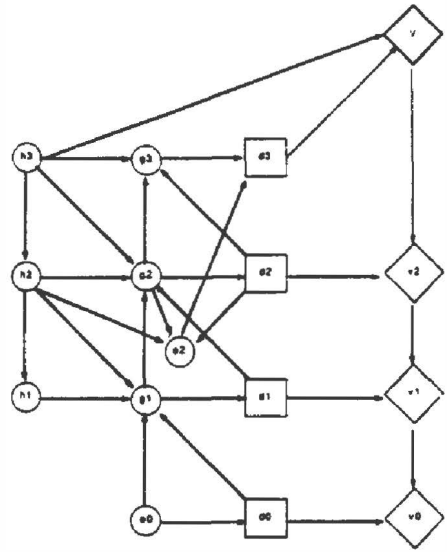

Figure 8: Refined Process Influence Diagram

This diagram violates Tatman's requirements. In particular $h_2$ is a predecessor of $g_2$,

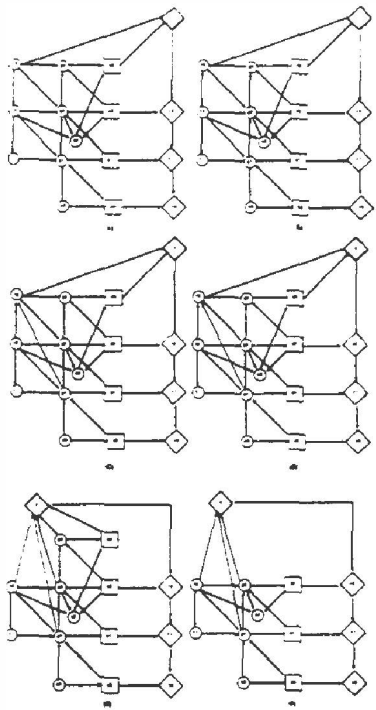
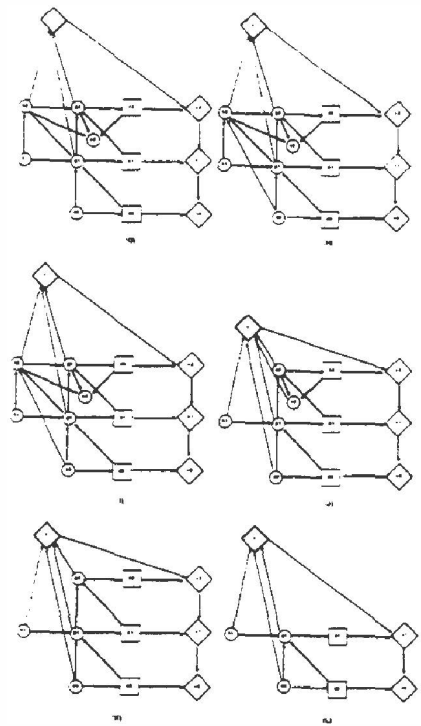

Figure 9: Solution Steps for Refinement Processing A-F

Figure 10: Solution Steps for Refinement Processing G-L

and $e_2$ is a successor of $g_2$, but there is a path from $h_2$ to $e_2$ that does not pass through $g_2$. Also, $e_2$ is not a predecessor of $d_2$, so it cannot be included in $X(2)$. (Notice that $e_3$ was a predecessor of $d_3$.)

However, by applying the same proof principles to this particular diagram structure that Tatman applied in his optimality proof, we can show that sub-value modularity is maintained, at the cost of complicating each stage with an additional arc. To see this we apply the standard influence diagram solution steps to roll back the diagram as shown in Figures 9 and 10.

As we reverse arcs and connect predecessors, the computational complexity rises from steps a-e, and then as nodes are absorbed, at the third level, the diagram simplifies almost back to the original, incremental two-stage (up to $h_2$) diagram, except that we have an extra arc from $g_2$ to the top value node. As we continue rolling back the diagram, reversal of arcs entering $e_2$ again raises complexity, and again resolves with only one additional arc from $g_1$ to the value node. We see that the incremental solution maintains the modularity of the original dynamic programming method so that the solution time is still linear in the number of stages.

## VI. Conclusions

We have formalized an approach to machine vision in an influence diagram framework and shown that system processing can be represented as dynamic instantiation of image interpretation hypotheses in influence diagrams. Hypotheses are generated by matching aggregated imagery features against physical object models. Instantiating new hypotheses correspond to introducing new nodes and random variables in the influence diagram. We showed a method of representing the affects of basic imagery interpretation actions of search, refinement and aggregation in influence diagram formalisms. Each new action that is taken leads to a new influence diagram. We chose the next vision action by evaluating the current diagram. The final influence diagram contains all random variables dynamically in-stantiated during control by the vision system. We showed that the sequence of decisions to act taken by the vision system is the same as it would have been had we derived those decisions from evaluation of the final influence diagram.

Our method of vision system control by incrementally evaluating influence diagrams as we build them results in a consistent, evaluated, final influence diagram. Development of an efficient evaluation method for partial instantiation of diagrams remains as future research.

There is much more work to do to complete the task of machine vision system representation and execution. So far we have only represented aggregation, search and refinement between neighboring hierarchical levels of inference. However, the general vision problem allows these operations to jump around between levels. This in turn raises the issue of classifying machine vision operations in terms of their probabilistic dependencies. We believe we have captured some fundamental paradigms for computer vision, but there are many operators and processing paradigms in the literature. For example aggregation operations between inferred 3D volumes and adjacent 2D surfaces involves violating modularity assumptions used in this work.

Another major issue is the pre-runtime computation of expected values from system processing. In [Levitt et al.-88] a scheme was presented for hierarchical value computation. This work shows that because we can cast machine vision control as a dynamic programming construct, the concept of value of information can be applied. Casting this concept in this framework is work in progress.

Finally, the combinatorics of machine vision demand distributed processing. This requires multiple processing decisions to be made simultaneously. Here optimality computation is burdened with the expected interactions between processing results. It is likely that many more engineering solutions will be realized before a formal analysis of this problem is completed.

## Acknowledgements

Partial support for this research has been provided under the ADRIES contract (U.S. Government Contract DACA76-86-C-0010) by the Defense Advanced Research Projects Agency (DARPA), the U.S. Army Engineer Topographic Laboratory (ETL), the U.S. Army Space Program Office (ASPO), and the U.S. Air Force Wright Avionics Development Center (WADC).